# Exploring Metaphorical Senses and Word Representations for Identifying Metonyms

**Wei Zhang**
Carnegie Mellon University
5000 Forbes Ave.
Pittsburgh, PA 15213
`wynnzh@gmail.com`

**Judith Gelernter**
Carnegie Mellon University
5000 Forbes Ave.
Pittsburgh, PA 15213
`gelern@cs.cmu.edu`

**Abstract**

A metonym is a word with a figurative meaning, similar to a metaphor. Because metonyms are closely related to metaphors, we apply features that are used successfully for metaphor recognition to the task of detecting metonyms. On the ACL SemEval 2007 Task 8 data with gold standard metonym annotations, our system achieved 86.45% accuracy on the location metonyms. Our code can be found on GitHub[1].

## 1 Introduction

*"Boston lost to Baltimore"*. The cities—in this sentence, the abbreviated names for sports teams—are metonyms, a type of figurative language where one word substitutes for a related word or phrase.

Disagreement shows that this is a difficult problem that automation will be unable to resolve fully. People disagreed about whether location words are literal or figurative (metonymic) about 15% of the time in the case of the official annotators for the SemEval metonym data set we used (Markert and Nissim, 2009), and about 8% of the time in our own assessment of the annotations on this standard data set.

The significance of recognizing location metonyms is demonstrated by its commonplace occurrence. 17% of all location names in a corpus of manually-tagged German newspapers were metonymic (Leveling and Hartrumpf, 2008), and the SemEval 2007 Task 8 data includes locations which are about 20% metonymic, to reflect the "in the wild" distribution. Leveling and Hartrumpf (2008) found that removing metonyms from an index of literal and metonymic location names improved the performance of geographic information retrieval from a mean average precision of .0857 to .0917; Leveling and Veiel (2007) also found that excluding metonyms improved precision in identifying place names. Location grounding (Zhang and Gelernter, 2015) could benefit from excluding location metonyms as well.

*Metonym vs. Metaphor*. Some linguists see metonyms and metaphors as quite different. "In Metonymy one entity stands for another, in Metaphor, one entity is viewed as another" (Fass, 1991). By contrast, Radden (2000) considers metonyms to be a particular type of metaphor. Radden's understanding allows us to use abstraction as a feature in helping to classify metonyms.

*Method.* For the purposes of our research, what matters is that both metaphor and metonymy are abstract comparisons of two things. That makes relevant for our work the metaphor detection methods that draw upon word extraction. Other metonym detection methods, by contrast, do not draw upon word abstraction.

Some of our features for machine learning are constrained by the data set, which we used in order to compare to other research teams. Our results reflect the sparseness of metonyms in the training data (that has location metonyms to physical locations at 1:5). Were we to collect data via machine learning that had a larger ratio of location metonyms to physical locations, we would be able to train a more robust model and achieve higher scores. Future work will depend on richer data to generalize the model.

*Evaluation*. We use the SemEval 2007 data set so that we can compare our results to others.

---

[1] https://github.com/weizh/metonym

Because of the algorithm tailoring necessary to compare our approach to others on this particular data set, we did not use additional data sets.

***Our contribution.*** (1) We created a metonym recognition system that uses abstraction, and (2) that explores various word representations that were never before published in metonym recognition experiments.

## 2   Related Work

Markert and Nissim (2002) were the first to propose that metonym resolution as a classification problem. Then Nissim and Markert (2003) proposed to use syntactic features and word similarity for the task.

The SemEval 2007 Task 8 provided metonym annotations that have been widely used for the problem.[2] We use the metonym *recognition*, from Named Entity Recognition, rather than metonym resolution as is used in the SemEval 2007 literature, because the metonyms are not resolved into anything, but only identified.

The SemEval 2007 data was taken from the British National Corpus, and it is the same data that we use for our experiment. Five groups competed, as chronicled by Markert and Nissim (2009). Overall, the systems were much better at identifying the literal sense of a location than at identifying location metonyms.

Unhelpful at identifying metonymy were shallow features such as part of speech. Features which proved more helpful were the subject/object role in the sentence, as seen by the relatively high accuracy of the baseline that marked the potential metonymic candidate as metonymic if it were a subject, and non-metonym if it were not a subject.

The winning system in the SemEval 2007 task used a maximum entropy learner with grammatical and lexical features for words, and between words (Farkas et al, 2007). Nastase and Strube (2009) achieved good results by combining lexical, encyclopedic and collocation information. Later, Nastase et al., (2012) added global context in an unsupervised method that rivals supervised methods. But after that, the task was studied rarely.

## 3   Method

Our task in metonym recognition is to define features that distinguish between metonymic and literal uses of a potential metonymic word for location. This section describes the abstractness and imaginability features we borrowed from prior research on metaphor detection (Tsvetkov, 2013 et al),[3] and our features for word representation, some of which had not been used before for this purpose.

We used the given data set so that we could compare to other methods. But then we used the designated "grammatically-related word" in each extract in the data set as a feature. Other groups, too, used this feature for metonym detection.

### 3.1   Intuition underlying the method

One assumption for our method is that a potentially metonymic word and the nearby grammatically related word (if not a preposition) share the same degree of abstractness. The examples below illustrate.

***Potentially metonymic word--* in a name phrase**
Location as part of a named entity phrase. e.g., "*U.S.* Federal trade commission". The potentially-metonymic word is "U.S.", and the grammatically-related word "commission" is a concrete word that indicates the literal sense. In contrast, in the phrase "U.S. position on global warming", the grammatically-related word "position" is an abstract word and an indicator for the metonymic sense for potentially-metonymic word "U.S.". This shows metonymy of a potentially metonymic word can be indicated by the abstractness of grammatically-related word.

***Potentially metonymic word--* Action** Locations used as "place for people", such as "Israel's absorption of Palestinians," where the grammatically-related word "absorption" is a strong indicator of an abstract action, confirming the metaphorical sense of the potentially-metonymic word. Note that it is also "imaginable," which means we can create a picture in the mind to imagine it. The distinction between "abstractness and "imaginability" might also help us better understand the metaphorical sense of a metonym.

---

[2] SemEval Task 8 specified at http://nlp.cs.swarthmore.edu/semeval/tasks/task08/summary.shtml

[3] Note that Tsvetkov et al use the term "imagability", which is not found in the dictionary.

**Preposition-Potentially Metonymic Word** e.g., "the spine of Norway". The grammatically-related word is "of"[4]. Further information is needed to determine the sense of the potential metonymic word "Norway". Thus, prepositions are not evaluated for abstractness, However, some location indicators that work well in named entity recognition, such as "to" or "in," are good literal indicators for a nearby potentially metonymic word.

**Verb-Potentially Metonymic Word** e.g. "provide Albania with..." where the grammatically-related word is "provide", indicating the potentially-metonymic word "Albania" to be a "place for people" sense. This is the case where verb sense or verb class can take effect as well.

In some cases above, abstractness aligns with the metonymic sense of the potentially-metonymic word.

### 3.2 Abstractness and imaginability (A&I)

We use a knowledge base lookup to identify which words are abstract or imaginable.

Abstractness, according to the Webster dictionary, means "relating to or involving general ideas or qualities rather than specific people, objects, or actions."[5] But sometimes an abstract concept is imaginable, meaning "that may be imagined."[6] Sometimes people can call up an image of a concept in brain even it is abstract. For example, "vengeance" calls up an emotional image, whereas "torture" calls up emotions as well as visual images (Tsvetkov et al., 2014). The difference between the two can help us understand better what an abstract sense really means. In metaphor detection, abstractness and imaginability were shown to be useful as conjunction features to denote the conflict in abstractness degree of words (Broadwell et al., 2013; Turney et al., 2011; Tsvetkov et al., 2014),.

---

[4] According to grammatical annotation provided by SemEval 2007 Task 8 data set.
[5] http://www.merriam-webster.com/
[6] http://www.webster-dictionary.org/

Given the similarity between metonym and metaphors, metonym resolution should also benefit from it. In metaphor detection, the conflict between the two words is evaluated, but in metonym detection, the abstractness of the related word makes it more likely that the metonym candidate is, in fact, metonymic.

### 3.3 Word representations and abstractions

A word representation often takes the form of a numerical vector associated with each word. Each dimension's value in the vector corresponds to a feature, and might even have a semantic or grammatical interpretation, so we call it a word feature (Turian et al, 2010).

In this research, word representations are a means to detect subtle semantic patterns of the metonym or non-metonym and the surrounding words in the sentence. Word representations essentially expand words in the data by using related words or vectors that carry semantic information, so as to improve inference potential.

We use several word representation methods here. That is because sometimes in Machine Learning, a method that does not improve results by itself, might improve results when combined with others. This interaction between word representation features is demonstrated in Table 2.

## 4 Features

### 4.1 Knowledge base for abstractness

The MRC psycholinguistic database is a dictionary with linguistic and psycholinguistic attributes that were obtained experimentally (Wilson, 1988). It includes 4295 words rated by degree of abstractness and 1156 words rated by degree of imaginability.

Our hypothesis is that the relative degree of concreteness or abstractness of the dependent word might indicate also the relative degree of concreteness or abstractness of the potentially-metonymic word. But we found that evaluating for abstractness does not indicate strongly whether the candidate location is a metonym or not.

Tsvetkov et al., (2014, 2013) trained a logistic regression classifier with the MRC words using the vector space representation as a feature. They propagate the abstractness and imaginability scores to all the words by classifying every word that has vector space representation.[7] They reported an accuracy of 0.94 and 0.85 respectively for abstractness and imaginability predictions of metaphor. The data is available on GitHub.[8]

We did not need to train a similar classifier for metonyms in the SemEval dataset because our words happened to appear among the word vectors created by the Tsvetkov et al. (2014). Note that the abstractness classifier would probably predict every potential metonym as concrete, so we use only the abstractness of the grammatically related word to indicate whether the potential metonymic word is indeed a metonym. But in metaphor detection, every word in the subject-verb-object and adjective-noun structure can be evaluated for abstractness, so disagreement on degree of abstractness could be generated to indicate metaphor.

### 4.2 Word Representation Feature

We employ two types of word representations: *word vector representations*, and *word abstraction* i.e. hyper senses and concepts. Both of them aim to handle the problem of sparsity in the training data.

*Vector representations*

Vector space word representation is a general term that includes one-hot coding and word embedding. Word embedding is a technique that maps words from the document vocabulary to real number vectors in a space of low dimensions relative to the size of that vocabulary.

Different vector representations have different advantages. We use three types:

(1) word embeddings, or distributed word representations (Turian et al., 2010), that capture the "semantics" of each word.

(2) distributional word representations such as Latent Semantic Analysis (Landauer et al.,1998), ICA (Väyrynen et al., 2007) and LDA (Blei et al., 2003) that capture the corpus-level word distribution, and

(3) word one-hot coding vectors (or bag-of-words vectors) which preserve word forms, which (1) and (2) fail to encode (Turian et al, 2010). The vector representations are complementary, and their combination could make better generalizations of grammatically-related words.

We used the 25 dimensional neural language model trained with single layer neural network as in (Turian, 2010) for the word embeddings.

We used the 64 dimensional vector trained with a variation of Latent Semantic Analysis by Faruqui and Dyer (2014). The vector construction algorithm is a variation of traditional Latent Semantic Analysis (Deerwester et al., 1990) to produce representations in which synonymous words have similar vectors.

We convert the grammatically-related word into a one-hot coding vector if the grammatically-related word is in a dictionary, otherwise it gets a vector of zero. One-hot coding creates a feature vector the length of the location word dictionary. The model does not generalize well to words not included in the labelled training data, however, because similarity among words is not encoded. That is why we need other word representations such as embeddings.

*Word Abstractions*

We use Brown clusters, Levin verb categories, and WordNet concepts to generalize the grammatically-related words to higher level abstractions.

***Brown Cluster.*** The Brown algorithm clusters words by maximizing mutual information and similarity within the corpus (Brown et al, 1992). We used clusters trained on CoNLL2003 [9] (Turian et al, 2010). Each word is generalized into a Brown Cluster ID and semantically equivalent word (such as Monday and Tuesday) or a synonymous word that has identical ID prefixes. We found through experimentation that a cluster ID of prefix length 4 is most effective.

***Levin*** has divided most commonly-used English verbs into categories, according to verb semantic usage.[10] These categories have been used in previous metonym research (Farkas et al., 2007). We collected all of the Levin categories by category ID for all the verbs in the training set. We used a hash map to store the relation of category and verb. Then to make features to represent the Levin categories, we found all the relevant IDs with the verb in the hash map. Then we made a vector of all zeros of the length equal to the category dictionary, and filled in the corresponding slots that are IDs found in the hashmap with ones.

---

[7] https://github.com/ytsvetko/metaphor/tree/master/resources

[8] https://github.com/ytsvetko/metaphor/tree/master/resources

[9] http://Metaoptimize.com/projects/wordreprs/

Note that in the Levin system, a verb can belong to more than one category, so we used N-hot-coding instead of one-hot coding to convert verb categories to vectors. This becomes a semantic vector that explicitly encodes the meaning of each slot, as distinct from word embedding vectors where each slot does not necessarily bear a specific meaning.

***WordNet Synonym set ID***.[11] So we can use the word and its part of speech to search the WordNet database. Part of speech is not provided for the SemEval 2007 data, so we simply took the first Synset ID of the word in WordNet, regardless of its part of speech. The syntactic head features works if the Synset of the word from the training data happens to be found in the test data. The Synset ID match thus will lessen the data sparsity problem.

Here we used the Synset ID of the WordNet entry, regardless of part of speech, as is also used by Farkas et al. (2007) and Nastase et al. (2009).

### 4.3 Miscellaneous features

We use some other features found effective as well (Nastase and Strube, 2009) (Farkas et al., 2007). Those features are the grammatical role (syntactic parse relation) of the location word as given in the *SemEval 2007 task 8* corpus annotation, and whether the dependent word is a preposition. Additionally, we add an indicator for whether the dependent word is in the stop word list.

### 4.4 Creating a metonym-verb list

We used Google Ngram to create a list of verbs

---
[10] http://www-personal.umich.edu/~jlawler/levin.verbs
[11] http://wordnet.princeton.edu/wordnet/download/

likely to indicate metonyms, but the list was ineffective. Here's what we did.

Google Ngram are part-of-speech tagged excerpts from books that show how words tend to be used in the language. We use word frequencies from Google Ngram. We were especially interested in how verbs have been used.

For each verb preceding or following a location word in the Ngram corpus, we created a list of how many times that verb was used with a person as the subject versus how many times that verb was used with a location as the subject. Then we divide the frequency of the verb used with a person by the frequency of the person word to generate a "person ratio" for the verb, and similarly, the frequency used with location verbs with the frequency of the location, to generate a "location ratio" for the verb, and then we compare ratios.

$$P(Verb \mid Person) = \frac{Count(PersonVerb)}{Count(Person)}$$

$$P(Verb \mid Location) = \frac{Count(LocVerb)}{Count(Location)}$$

If the person as subject vastly out-weighs the thing as subject, then we have discovered what linguists refer to as a "selective restriction" on the verb, and the candidate location is probably a metonym.

Our method was unsuccessful because Ngram does not give sentences, and therefore we had no sentence context to differentiate between Person and location verbs – we only had subject—predicate—object structure. Also, we did not have a comprehensive list of personal names and locations to get reliable counts for the equations.

## 5 Experiments, Results and Limitations

*Experiment data.* SemEval 2007 Task 8 location data corpus uses text from British National Corpus (BNC). The data is taken from a variety of sources, including newspapers, magazines, and transcripts of conversations.

Each instance in the SemEval 2007 Task 8 data includes about 2-4 sentences context for a single potential metonymic word annotations for literal or metonymic, and grammatically related word(s) (Markert and Nissim 2007). Also provided is tokenization and part of speech annotations for each sample.

*Ground truth annotation.* The location training set has 173 metonyms, and 737 literals, and the testing set has 167 metonyms, and 721 literals, all of which are country names. Annotations also contain the metonym sub-types Place-for-event, Place-for-people, Place-for-product, Object-for-name, and Other. Any ambiguous senses when the annotators disagrees were removed from training and testing data (Markert and Nissim 2007). Even so, as mentioned above, we disagreed with the gold annotations about 8% of the time in our own assessment.

Following is a randomly-selected extract from the training corpus for location metonyms: Only "Britain" (boldface in the example) is marked as a metonym in the gold standard, but there are far more literal and metonymic examples in a single excerpt, including Paris, Washington, U.S., etc.

> *Written by MIT Professor Paul Krugman, formerly on the staff of the White House Council of Economic Advisers, and Professor Edward Graham, formerly in the office of international investment at the US Treasury, and later at the OECD in Paris, the report will be scanned carefully as indicating the current thinking of the Washington policy-making establishment.*
>
> *"Much public attention has been focused on foreign purchases of real estate, however, foreign ownership of banking and manufacturing is much more important," the report argues.*
>
> *And while the bulk of public controversy has focused on Japanese investment,* **Britain** *plays by far the biggest role, accounting for 31 per cent of all foreign investment in the US, as much as the next two investors, the Dutch (14.9 per cent) and the Japanese (16.1 per cent) combined. Britain's eminence is likely to be brief.*

*Given annotation:*[12]
*Britain play (subject)*
*Place-for-people (metonym)*

*Method.* The machine learning algorithm needs to work well on a small number of training instances, be error tolerant, and handle sparsely encoded feature vectors without information loss. An excellent but classic algorithm such as the Support Vector Machine (Cortes & Vapnik, 1995) that handles sparsity surpasses the others on the SemEval 2007 data set. Tsvetkov et al. (2014) used Random Forest (Liaw and Wiener, 2002) for metaphor detection. We tested Random Forest, Lasso Regression and Sparse Coding, and they all showed results to SVM. Note that the data does not contain a development set, so over-fitting is possible. As for the SVM parameters, C is set to 1 to tolerate errors; the RBF kernel is used due to the data non-linearity, and gamma is set to 1/k, where k is the number of feature values. The feature vector is just the concatenation of feature vectors for each feature category mentioned in previous sections.

*Results.* Table 1 compares our results to the best performing systems using the F1 statistic, as was used in the SemEval 2007 competition. Omitted from Table 1 are the comparison baselines that mark every potential metonym as literal, at 79.4% accuracy, and the baseline that marks the potential metonym as literal if is the subject of the sentence, at 83% accuracy.

For the coarse-grained evaluation, our algorithm out-performs the baselines, and as shown in Table 1, is comparable to the formerly highest-scoring result from the SemEval 2007 (Farkas et al, 2007), and the later result on the same data (Nastase et al. 2009). The SemEval data also included annotations for medium and fine-grained, but classifying to this level is not our objective, so we do not report these results.

---

[12] We used the annotations of verb and metonym type as features for the classifier.

We show in Table 2 the results of feature combinations, and in particular, the effectiveness of word embeddings and word abstractness and imaginability. Word Embedding shows a 0.1% improvement comparing experiment 3 and 4; Abstractness and Imaginability show an additional 0.2% if we compare experiment round 4 and 6 (see Table 1)

Table 1 Breakdown of F1 by category. Note: Literal, Non-literal (**NonLit**), Metonym, Mixed, Other metonyms (**Other**), Place-for-people (**PFP**), and Place-for-event (**PFE**).

|  |  | Farkas07 | Nastase09 | Ours |
|---|---|---|---|---|
| **Coarse** | Literal | 91.2 | 91.6 | **91.9** |
|  | NonLit. | 57.6 | 59.1 | **59.4** |

Table 2 F1 for feature combinations. OH (*GRW* One Hot Coding), DP (*PMW* and *GRW* dependency), BC (Brown clustering), LV (Levin Verb), PR (is preposition), WN (WordNet concepts), EM (word embedding), VS (vector space representation), A&I (Abstractness and Imaginability)

| # | Feature Combinations | F1 |
|---|---|---|
| 1 | OH, DP, BC | 85.8 |
| 2 | OH, DP, BC, LV | 85.9 |
| 3 | OH, DP, BC, LV, PR, WN | 85.9 |
| 4 | OH, DP, BC, LV, PR, WN, EM | 86 |
| 5 | OH, DP, BC, LV, PR, WN, EM, VS | 85.6 |
| 6 | OH, DP, BC, LV, PR, WN, EM, A&I | 86.2 |
| 7 | OH, DP, BC, LV, PR, WN, EM, A&I, VS | 86.5 |

*Limitations*. The trained model is limited because the grammatically-related words in the data set were selected randomly, and manually.

## 6 Promising approaches to improve metonym detection

*More data*. The greatest obstacle to our machine learning algorithm is insufficient data. The SemEval data is tagged sparsely, and many more instances of literal and also figurative locations are found in the data than are annotated. A more thorough tagging of both the training and testing data could produce better results via machine learning.

Furthermore, although the proportion of metonyms found in the data echoes what would be found "in the wild", the imbalance of only about 20% metonym instances in the corpus impairs the ability of the machine learning algorithm to recognize metonyms. More tagged instances of metonyms would improve recognition accuracy.

*Domain*. Context of the paragraph helps determine whether the words used in the paragraph are likely to be used literally or abstractly. Examples of known domains which include a higher than average proportion of location metonyms are sports and politics.

*Syntax*. We could add more heuristics based on the relationship of the words in the sentence. For example, if a location is the first word in a two- or three-word noun phrase, then the location is literal because it is working as an adjective. We had difficulty creating similar heuristics to indicate location metonyms. However, heuristics that indicate non-metonyms would be also useful. Our experiments suggest that coding some of these rules into a stand-alone system, while including some of the machine learning-based features, would recognize metonyms better than our current model. This is what Levin et al (2014) proposed for the detection of conventional metaphors.

*Context* By examining the context of the verb, we can find the best WordNet Synset Future work then would refer to the semantics of verbs using FrameNet in attempt to find semantic restrictions on usage. For example, person-indicating verbs, such as "is supposed to", "want" "wish" might indicate a figurative use of a location word (Berlin wants, for example). We extracted such a list of person-indicating through use of Google NGram as described above but the results were not successful.

## 7 Conclusion

We applied the metaphor insight of Tsvetkov et al (2014) to abstractness and imaginability in order to identify *metonymic* senses of a word. Their method worked well, principally because metonyms are abstractions, just as metaphors.

Metonymic location words occur in sentences in ways very similar to how literal locations occur. It is thus subtlety of meaning that guides distinguishing metonymic from literal meaning. And it is the addition of abstractness features and other semantic word representations that make our method effective. This paper shows a new semantic perspective on solving metonym identification problem that might be expanded to other figures of speech that are abstractions.

**Acknowledgements**

This research was supported in part by the Intelligence Advanced Research Projects Agency (IARPA) via Department of Defense U.S. Army Research Laboratory contract number W911NF-12-C-0020. The U.S. Government is authorized to reproduce and distribute reprints for Governmental purposes notwithstanding any copyright annotation thereon. The views and conclusions contained herein are those of the authors and should not be interpreted necessarily as representing the official policies or endorsements, either expressed or implied, of IARPA, DoD/ARL, or the U.S. Government.